\documentclass[10pt,twocolumn,letterpaper]{article}

\usepackage{cvpr}              
\definecolor{cvprblue}{rgb}{0.21,0.49,0.74}
\usepackage[pagebackref,breaklinks,colorlinks,allcolors=cvprblue, linkcolor=red]{hyperref}

\title{Thermally Activated Dual-Modal Adversarial Clothing \\
against AI Surveillance Systems}

\author{Jiahuan Long$^{1, 2, 3}$ ~~~~~~ Tingsong Jiang$^{1,3*}$ ~~~~~~ Hanqing Liu$^{1,2,3}$ ~~~~~~Chao Ma$^{2}$ \\
~~~~~~Weien Zhou$^{1,3}$ 
~~~~~~Yang Yang$^{1,3}$
~~~~~~Wen Yao$^{1,3}$\thanks{Corresponding authors}\\
$^{1}$Defense Innovation Institute, Chinese Academy of Military Science~~~~ \\$^{2}$MoE Key Lab of Artificial Intelligence, AI Institute, Shanghai Jiao Tong University  \\
$^{3}$Intelligent Game and Decision Laboratory \\
{\tt\small jiahuanlong@sjtu.edu.cn, tingsong@pku.edu.cn, wendy0782@126.com}
}

\begin{document}
\maketitle
\begin{abstract}
Adversarial patches have emerged as a popular privacy-preserving approach for resisting AI-driven surveillance systems. However, their conspicuous appearance makes them difficult to deploy in real-world scenarios. In this paper, we propose a thermally activated adversarial wearable designed to ensure adaptability and effectiveness in complex real-world environments. The system integrates thermochromic dyes with flexible heating units to induce visually dynamic adversarial patterns on clothing surfaces. In its default state, the clothing appears as an ordinary black T-shirt. Upon heating via an embedded thermal unit, hidden adversarial patterns on the fabric are activated, allowing the wearer to effectively evade detection across both visible and infrared modalities. Physical experiments demonstrate that the adversarial wearable achieves rapid texture activation within 50 seconds and maintains an adversarial success rate above 80\% across diverse real-world surveillance environments. This work demonstrates a new pathway toward physically grounded, user-controllable anti-AI systems, highlighting the growing importance of proactive adversarial techniques for privacy protection in the age of ubiquitous AI surveillance.
\end{abstract}

\section{Introduction}
\begin{figure}
\centering
\includegraphics[width=0.9 \linewidth]{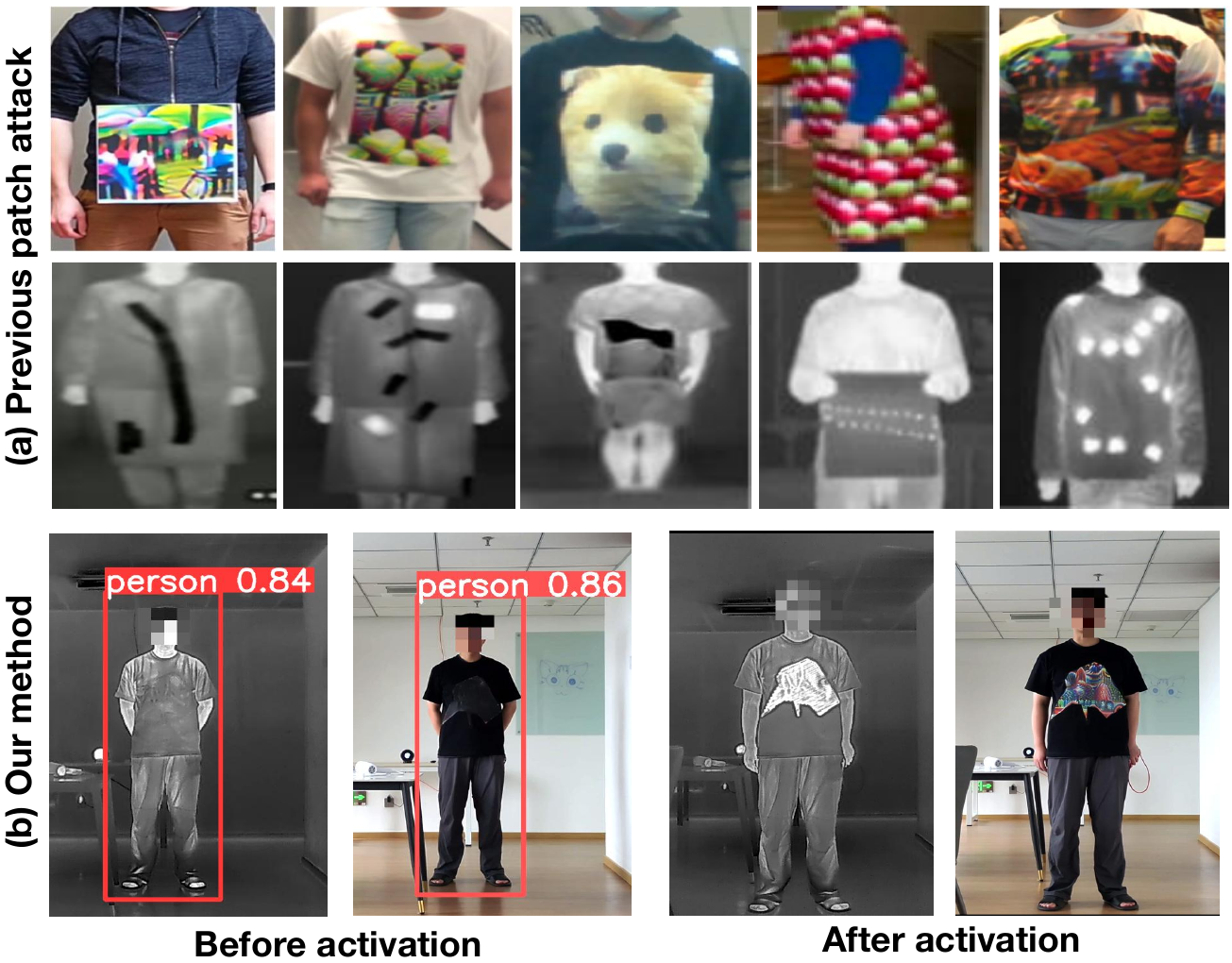}
\vspace{-0.1in}
\caption{\textbf{Comparisons of representative adversarial patch attacks.} (a) shows that most prior adversarial patches are single-modal (RGB or infrared) and always-on in everyday settings, making them more noticeable in real-world scenarios. (b) presents that our adversarial patch attack achieves dual-modal visible-infrared deception, and support controllable activation in the real world.}
\label{fig:intro}
\end{figure}
AI-driven surveillance systems have been widely deployed across human-centered public spaces, including schools, commercial facilities, and urban roadways~\cite{1-hussain2024ai, 2-kalluri2025computer, 3-zuboff2023surveillance, 4-maslej2025artificial, 5-feldstein2019global}. Such systems typically consist of visible and infrared cameras, edge/cloud computing units, and computer vision algorithms, enabling automated detection, classification, and tracking of targets (e.g., pedestrians, vehicles). Captured video streams are transmitted in real time to centralized platforms for automated decision-making or human-in-the-loop analysis~\cite{6-ardabili2023understanding}. According to the AI Global Surveillance Index~\cite{5-feldstein2019global}, AI-based surveillance has already been adopted by approximately 78 countries for applications such as smart city monitoring and predictive policing. By 2030, it is expected that more than 60\% of law enforcement agencies worldwide will integrate AI with live video feeds for public security monitoring.

As AI surveillance systems increasingly permeate daily life, their large-scale deployment in public spaces is raising unprecedented concerns over ethics and privacy security~\cite{2-kalluri2025computer, 8-browne2015dark, 9-chang2022countermeasures, 11-waelen2024ethics}. On the one hand, training these systems often relies on vast amounts of biometric data—including facial features, gait patterns, and behavioral trajectories—collected from individuals without explicit consent~\cite{12-van2020ethical, 13-perrigo2022clearview}. On the other hand, the lack of robust regulation and transparent data governance exposes users to severe risks of data misuse~\cite{14-barkane2022questioning}. According to the Stanford AI Index 2025~\cite{4-maslej2025artificial}, global AI-related privacy and security incidents increased by 56\% in the past year, including data breaches, model abuse, and identity tracking—leading to a significant decline in public trust regarding how AI companies and governments manage personal information.
These issues have intensified societal demand for proactive privacy protection, prompting growing public interest in anti-AI surveillance technologies.

Among anti-AI technologies, adversarial patch is a representative proactive countermeasures against AI surveillance~\cite{19-hingun2023reap, 20-hu2025two, 21-wei2023unified, 22-long2025cdupatch, 23-brown2017adversarial, 24-thys2019fooling, 25-wei2024revisiting, 26-liu2018dpatch, 27-long2024papmot, liu2025eva}. The patch is carefully designed visual pattern can be physically attached to clothing or belongings, causing AI models to misclassify or entirely overlook the wearer in the real world. In recent years, privacy advocates have increasingly adopted adversarial patches as a form of anti-surveillance fashion, integrating them into wearable clothing to safeguard personal data in public spaces~\cite{28-xu2020adversarial, 49-adversarialfashion, 50-capable2025clothing, hu2023physically}. However, existing adversarial patches are often high-saturation, high-contrast textures that maximize feature perturbations. While effective against AI detectors, such conspicuous patterns easily attract human attention, making them impractical for everyday social environments. Therefore, balancing attack effectiveness with real-world usability has emerged as a key challenge in adversarial patch design.

In this paper, we design a thermally activated adversarial clothing capable of on-demand activation and dual-modal deception. The garment integrates thermochromic dyes with flexible heating units to induce visually dynamic adversarial patches on the fabric. At ambient temperatures, it is indistinguishable from a standard black T-shirt. Upon local heating, the hidden adversarial pattern emerges, degrading visible- and infrared-spectrum detection under AI surveillance. As shown in Figure~\ref{fig:intro}, the thermal changes interfere with the infrared-spectrum detectors, while color changes mislead visible-spectrum detectors. This process can be rapidly activated within 50 seconds, allowing users to proactively protect their privacy when entering surveillance zones. Experiments across indoor and outdoor scenarios demonstrate that our adversarial clothing achieves strong performance and practical usability.




\section{Related Work}
\subsection{AI surveillance and privacy}
To prevent the misuse of AI-collected human data, the European Union enacted the AI Act (Regulation (EU) 2024/1689) in August 2024—the world’s first comprehensive legislation regulating artificial intelligence~\cite{15-eu2024aiact, 16-bird2025aiguide, 17-eprs2024legislation}. The law explicitly prohibits the deployment of real-time AI surveillance systems for biometric recognition in public spaces, except in cases of serious criminal investigations or terrorism threats. In the United States, 13 states have passed legislation containing provisions for AI data governance and privacy~\cite{18-davis2024stateai}. However, outside a few jurisdictions, most countries and regions still lack effective governance frameworks for AI surveillance technologies. Widespread issues such as data misuse, algorithmic opacity, and lack of accountability persist. This global regulatory vacuum has fueled growing public interest in anti-AI surveillance technologies.

\subsection{Adversarial patch attacks}
Previous digital attacks~\cite{jia2021iou, jia2022exploring, liu2025lighting, long2025robust, huang2024towards} are challenging to deploy in the physical world, as their perturbations are often imperceptible to the human eye and difficult to faithfully reproduce. 
In contrast, adversarial patches have emerged as a widely adopted physical attack method against AI systems~\cite{19-hingun2023reap, 20-hu2025two, 21-wei2023unified, 22-long2025cdupatch, 23-brown2017adversarial, 24-thys2019fooling, 25-wei2024revisiting, 26-liu2018dpatch, 27-long2024papmot, zhu2024infrared, zhu2025physical, guo2025physpatch, hingun2023reap, chen2022shape, zheng2024physical, zheng2025revisiting, wei2022adversarial, wei2024infrared, wang2021dual, wang2021universal, liu2019perceptual, liu2020bias}. Early patch works focused  on single-modal attacks. Thys et al.~\cite{24-thys2019fooling} demonstrate physically realizable adversarial patches that successfully mislead person detectors in the visible domain, while Zhu et al.~\cite{zhu2024infrared} propose infrared adversarial car stickers generated via a 3D mesh–shadow optimization to deceive thermal detectors. More recently, several works have extended patch attacks to visible–infrared dual-modal detectors~\cite{20-hu2025two, 22-long2025cdupatch, 21-wei2023unified}. For example, Long et al.~\cite{22-long2025cdupatch} propose CDUPatch, a color-driven cross-modal adversarial patch that achieves unified visible-infrared attacks by mapping RGB colors to corresponding thermal responses. Wei et al.~\cite{21-wei2023unified} propose a unified adversarial patch that simultaneously fools dual-modal detectors by optimizing patch shape with boundary-limited shape optimization, demonstrating high cross-modal physical-world attack success.
However, existing patch attacks are always-on in everyday settings, which increases their conspicuity and complicates real-world deployment. In this paper, we co-design a thermally activated patch to mislead both visible- and infrared-spectrum detectors while remaining concealed within ordinary clothing. By leveraging temperature as a control signal, our system activates both the visible and infrared adversarial patterns simultaneously, enabling controllable, on-demand dual-modal evasion.

\begin{figure*}[t]  
\centering
\includegraphics[width=1.0\linewidth]{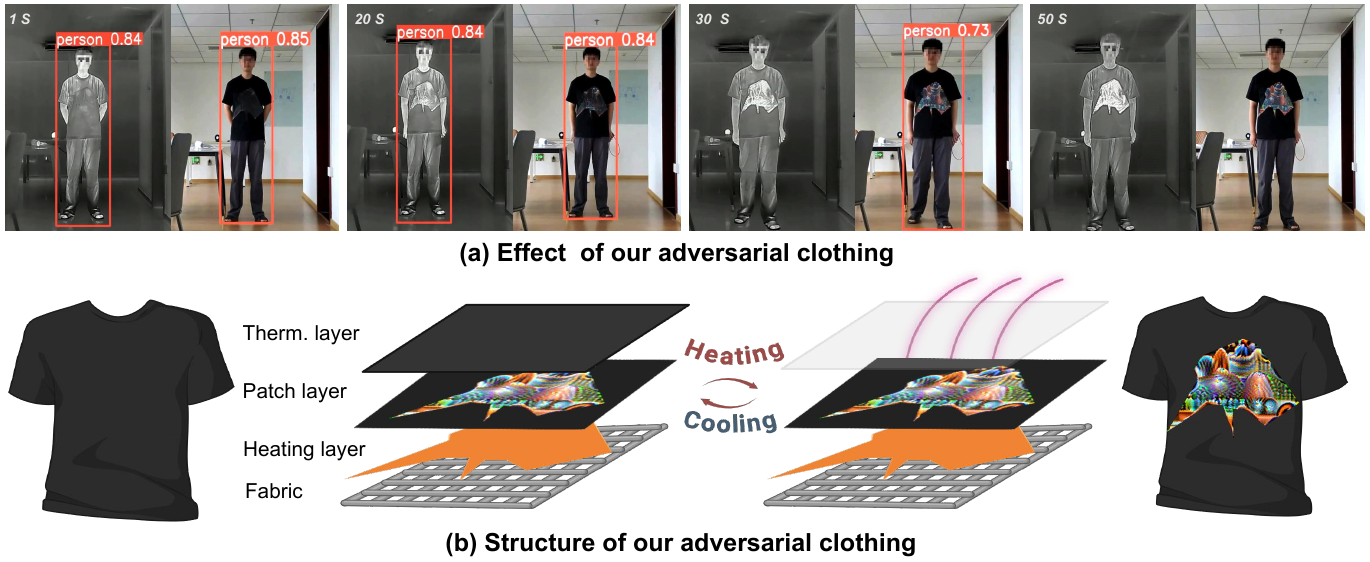}
\caption{\textbf{Effect and structure of the thermally activated adversarial clothing.} (a) Before activation, the wearer is detected by visible- and infrared-spectrum detectors; after activation, the adversarial pattern emerges and degrades the detection. (b) Layered structure of the clothing (top to bottom): thermochromic layer, adversarial patch layer, heating layer, and fabric substrate. When the heating layer raises the temperature above 30 ºC, the thermochromic layer becomes transparent, revealing the hidden adversarial patch; cooling restores the original black appearance.}
\label{Fig:Effect and structure of the adversarial clothing}
\end{figure*}

\section{Methodology}
Our design aims to realize a thermally activated dual-modal adversarial clothing system capable of providing on-demand deception in both the visible and infrared spectra. In this section, we first formalize the attack objective, then describe the structure of adversarial clothing, the thermal activation mechanism, and the algorithm used to optimize the cross-modal adversarial patch.

\subsection{Objective function}
Let $\mathcal{D}=\{(I^{v}_{j},I^{r}_{j})\}_{j=1}^{N}$ be a dataset of dual-modal image pairs, where each sample consists of a visible image $I^{v}_{j}$ and its aligned infrared image $I^{r}_{j}$. Our objective is to optimize a single adversarial patch $p$ that, when simultaneously applied to both $I^{v}_{j}$ and $I^{r}_{j}$, can fool the visible--infrared detector.


Let $T$ denote the visible--infrared object detectors with parameters $\theta$. The patch is optimized to jointly reduce detection confidence in both visible and infrared domains:
\begin{equation}
    p \;=\; \arg\min_{\kappa}\;
    \mathcal{L}_{\mathrm{adv}}\!\left(
        T\!\left(I^{v}_{j},I^{r}_{j}\mid \theta\right),
        \mathbb{D}_{j},\kappa
    \right),
\end{equation}
where $\kappa$ denotes the pixel values of $p$, $\mathbb{D}_{j}$ is the set of detection outputs (or ground truth) for image pair $j$, and $\mathcal{L}_{\mathrm{adv}}$ is the adversarial loss function.

\subsection{Structure of adversarial clothing}
To enable proactive privacy protection under AI surveillance, we design an adversarial clothing featuring on-demand activation and dual-modal visible-infrared camouflage. As shown in Figure~\ref{Fig:Effect and structure of the adversarial clothing}(a), before activation the person is detected by infrared- and visible-spectrum detectors with high confidence (0.84 and 0.85, respectively); after activation, both detectors fail to identify the person (confidence $<$ 0.4). During this process, thermal contrast perturbs the infrared-spectrum detector, while color change deceives the visible-spectrum detector (see Supplementary Materials for the video demoxx). Figure~\ref{Fig:Effect and structure of the adversarial clothing}(b) illustrates the layered structure of the clothing: (i) a \textit{thermochromic layer} that switches from black to colorless when heated ($\sim$30 ºC), (ii) an \textit{adversarial patch layer} carrying the algorithmically optimized RGB texture, (iii) a \textit{flexible silicone heating layer} with precise temperature control, which activates the thermochromic layer and generates controlled infrared patterns, and (iv) a \textit{fabric substrate} that provides thermal insulation for the wearer. Detailed descriptions of each layer are provided in the following subsections.

\subsection{Thermochromic layer}
\begin{figure}[t]  
\centering
\includegraphics[width=1.0\linewidth]{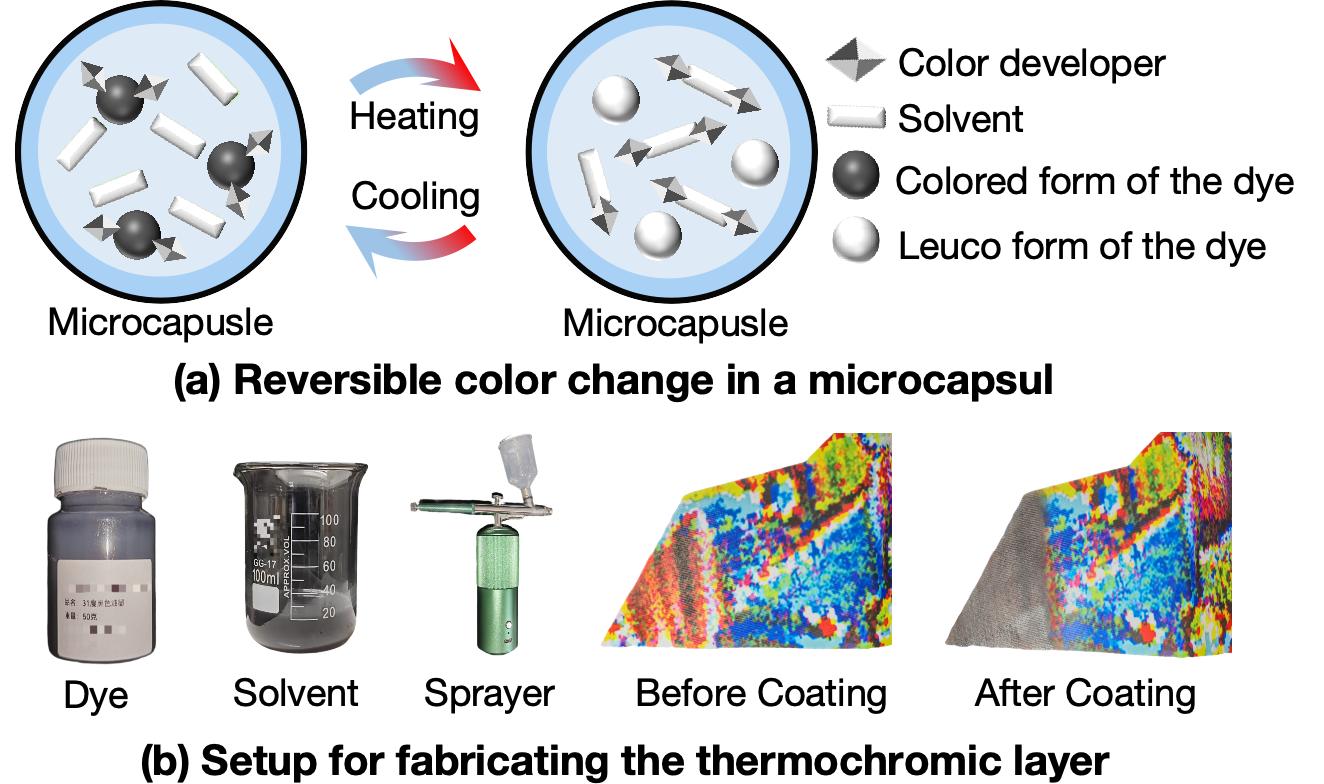}
\caption{\textbf{Characterization and working principle of microcapsule-based thermochromic dyes.} (a) Schematic illustration of the reversible color change in a microcapsule. At low temperature the dye remains colored, while heating switches it to a leuco (colorless) form. (b) Setup for fabricating the thermochromic layer: microencapsulated thermochromic dye, cyclohexanone solvent, airbrush sprayer, and the textile samples before/after coating.}
\label{Fig:microcapsule}
\end{figure}

The thermochromic layer is fabricated as a thin dye coating that undergoes a reversible transition from black to colorless upon heating. 
This layer consists of thermochromic microcapsules dispersed in the coating matrix, and its color-changing property originates from these microcapsules~\cite{51-liu2020reversible,52-liu2023high,54-tang2024electrothermochromic}. 
Figure~\ref{Fig:microcapsule}(a) illustrates the reversible color-change mechanism of a microcapsule. 
Each microcapsule contains a color developer (electron donor), a leuco dye (electron acceptor), a solvent, and a resin shell. 
When the temperature is below the solvent's melting point, the solvent remains solid, enabling a $\pi$-conjugated system to form between the color developer and the leuco dye, which causes the microcapsules to appear colored. 
Once the temperature exceeds the solvent's melting point, the solvent transitions to liquid form, disrupting the $\pi$-conjugated system. 
At this stage, the microcapsules do not display color and instead appear nearly transparent. 
Based on the above principles, the color-change temperature of microcapsules can be controlled by adjusting the composition of the solvent. 
In this study, microcapsules with a color-change temperature of around $30^{\circ}\mathrm{C}$ were selected. \textbf{It is worth noting that the activation temperature can be tuned by adjusting the solvent composition of the microcapsules.}

Figure~\ref{Fig:microcapsule}(b) presents the setup for fabricating the thermochromic layer. Specifically, 5 mL of thermochromic dye was diluted with cyclohexanone in a 1:3 ratio, ensuring uniform mixing by stirring, followed by ultrasonic treatment for 10 seconds. The resulting solution was then transferred to an airbrush container and sprayed onto the textile substrate. After coating, the underlying textile pattern is masked by a dark thermochromic overlayer, which becomes nearly transparent upon heating. The thermochromic dye was sourced from Quanzhou Hongchen New Materials Co., Ltd.

\subsection{Heating layer}
\begin{figure}[th]  
\centering
\includegraphics[width=1.0\linewidth]{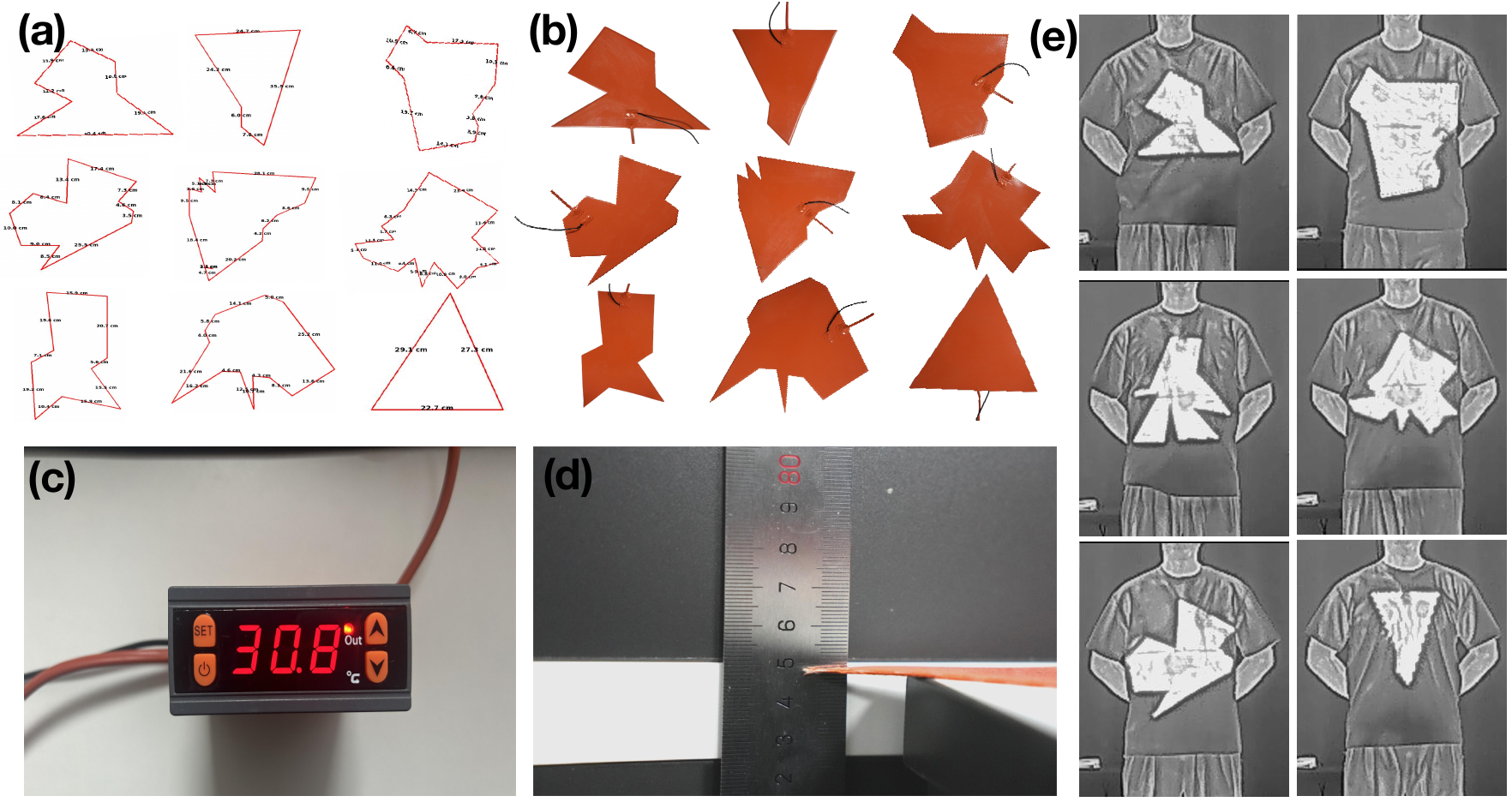}
\caption{\textbf{Design of the temperature-controlled heating pad.} (a) Algorithmically optimized polygonal shapes. (b) Silicone pads fabricated according to these shapes. (c) Digital temperature controller. (d) Thickness of the heating pad ($\sim$1 mm); (e) Visual examples of heating pads under infrared imaging.}
\label{Fig:heating pad}
\end{figure}
To ensure stable thermal conditions for activating the adversarial clothing, we integrated flexible, silicone-based heating pads into the textiles. Figure~\ref{Fig:heating pad}(a) presents the various pad shapes optimized using our adversarial patch optimization algorithm. Heating pads fabricated according to these shapes generate corresponding adversarial textures in infrared imaging. 
Figure~\ref{Fig:heating pad}(b) shows the heating pads based on these optimized shapes. Inside each pad, the heating element consists of nickel-alloy wire wound around fiberglass, ensuring uniform heat distribution with a power output of up to 0.4~W/cm$^{2}$. For safety in practical applications, the pads are encased in a composite insulating layer of silicone rubber, which provides excellent insulation with a breakdown voltage of 20--50~kV/mm. Each pad is powered by a portable power bank and regulated by a digital temperature controller, offering a controllable temperature range of 20--70~$^{\circ}$C (see Figure~\ref{Fig:heating pad}(c)). The pads have an approximate thickness of $\sim$1~mm (see Figure~\ref{Fig:heating pad}(d)), providing flexibility for integration into the fabric. When activated at 30~$^{\circ}$C, the pads produce distinct infrared patterns against infrared-spectrum detectors, as shown in the visual examples in Figure~\ref{Fig:heating pad}(e).


\subsection{Adversarial patch layer}
\begin{figure*}[t]  
\centering
\includegraphics[width=1.0\linewidth]{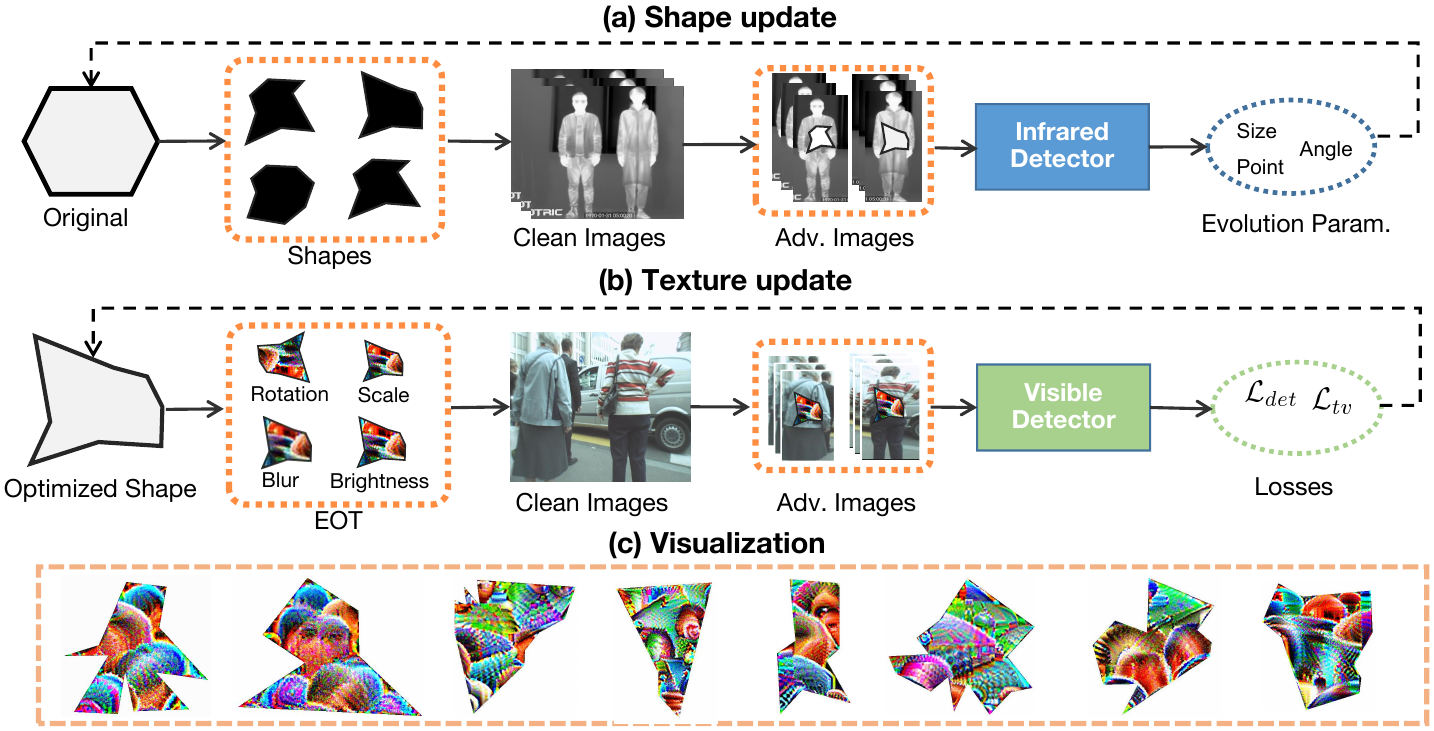}
\caption{\textbf{Overview of the proposed dual-modal patch training framework.} (a) Shape update against infrared detector. Starting from an initial geometric template, we optimize the patch shape to evade the infrared-spectrum detector by adjusting the number of vertices and the polar coordinates (radius, angle). (b) Texture update against an visible-spectrum detector. Given the optimized shape, we further update the patch texture under EOT transformations to fool the visible-spectrum detector by minimizing adversarial losses. (c) Representative adversarial patches obtained after the two-stage optimization.}
\label{Fig:framework}
\end{figure*}
The adversarial patch layer carries the algorithmically optimized RGB texture against visible-spectrum detectors. When the heating pad raises the temperature, the thermochromic layer becomes transparent, revealing the underlying adversarial pattern. The core objective of our method is to optimize a cross-modal adversarial patch that maintains strong attack performance against visible-infrared detectors in AI-surveillance scenarios. The optimization framework of patch comprises two key phases: shape update and texture update. 

In the shape update phase (Figure~\ref{Fig:framework}(a)), we optimize the patch geometry to evade infrared detection. Starting from an initial polygonal pattern, we iteratively evolve its shape by adjusting parameters such as vertices, radius, and angle. Each candidate shape is pasted onto every person instance in the infrared images to generate adversarial examples, which are then evaluated by an infrared detector. An attack is considered successful if the patched person is not detected (confidence $<$ 0.5). We then compute a person-level attack success rate (ASR) across the dataset and select the shape with the highest ASR to the next stage. 

In the texture update phase (Figure~\ref{Fig:framework}b), we fix the optimized shape and learn a color texture that deceives the visible-spectrum detector. To account for real-world variations, we adopt the Expectation over Transformation~\cite{48-athalye2018synthesizing} (EOT) to simulate real-world transformations (e.g., rotation, brightness changes, blur, and scaling), thereby improving the patch’s attack effectiveness under diverse physical conditions. Then, the transformed patch is directly pasted onto RGB images to form visible-spectrum adversarial images. Finally, these patched images are fed into the visible detector, and the patch texture is updated via adversarial losses. We adopt a combination of total variation loss ($\mathcal{L}_{tv}$) and average precision loss ($\mathcal{L}_{ap}$) as the final adversarial objective, which is widely used in adversarial patch optimization~\cite{24-thys2019fooling, 22-long2025cdupatch, 38-wu2020making, 30-hu2022adversarial}.  

Since the search landscape is highly non-convex, shape optimization often produces multiple competitive candidates. To preserve diversity, we retain a top-K set of diverse shapes for further evaluation. Representative optimized patches obtained after the two-stage process are shown in Figure~\ref{Fig:framework}(c).

\begin{figure*}[t]
\centering
\includegraphics[width=1\linewidth]{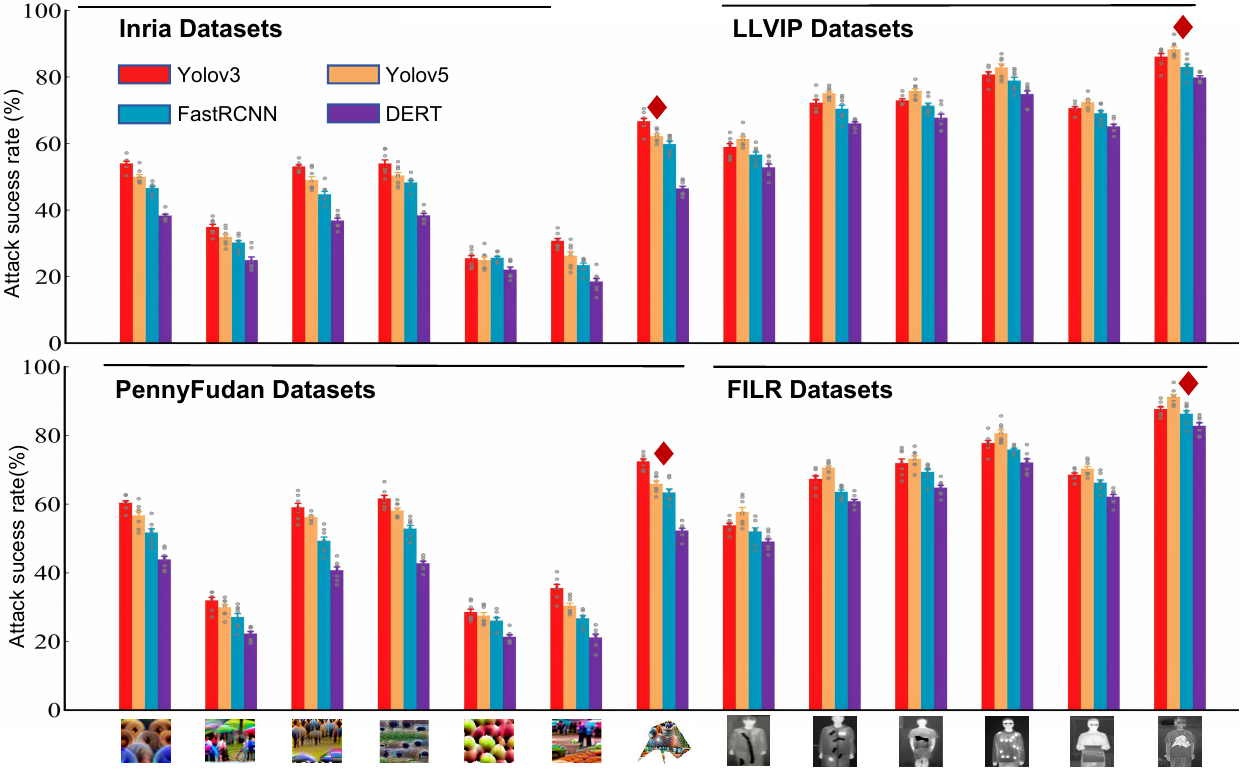}
\caption{\textbf{Comparison of attack success rates (ASR) for various adversarial patch methods on infrared and visible datasets.} Columns from left to right: T-SEA~\cite{36-huang2023t}, AdvYOLO~\cite{24-thys2019fooling}, CAP~\cite{25-wei2024revisiting}, FDA~\cite{37-cheng2024full}, AdvTexture~\cite{30-hu2022adversarial}, AdvCloak~\cite{38-wu2020making}, \textbf{our RGB patch}, AdvIC~\cite{31-hu2024adversarial}, AdvIB~\cite{32-hu2024adversarial}, AIP~\cite{33-wei2023physically}, HIC-IR~\cite{34-zhu2023hiding}, Bulb~\cite{35-zhu2024hiding}, \textbf{our infrared patch}. It shows that our dual-modal patch achieves the highest attack success rate compared to other representative patch methods.}
 \label{fig:data}
\end{figure*}

\section{Experiments}

\subsection{Experimental Settings}

{\noindent\bf{Datasets and victim models.}} For digital attack experiments, we evaluate our adversarial patch on person detection benchmarks using the FLIR~\cite{40-flir_dataset}, INRIA~\cite{41-dalal2005histograms}, PennyFudan~\cite{42-wang2007object}, LLVIP~\cite{39-jia2021llvip} datasets. The FLIR and LLVIP dataset are used to assess attack performance under infrared spectrum detection, while the INRIA and PennFudan datasets are used for evaluation in the visible spectrum. 
We evaluate our attack on three mainstream detection paradigms: one-stage detectors (i.e., YOLOv3~\cite{43-redmon2018yolov3}, YOLOv5~\cite{44-jocher2020yolov5}), two-stage detectors (i.e., Faster R-CNN~\cite{45-ren2015faster}) and transformer-based detectors  (i.e., DETR~\cite{46-carion2020end}). To ensure accurate detection of both infrared- and visible-spectrum pedestrian targets, we use the officially pre-trained weights using the COCO~\cite{47-lin2014microsoft} dataset as the initial weights and then fine-tune these models on the training sets of INRIA and FLIR datasets. After fine-tuning, all detection models achieve over 90\% mean Average Precision (mAP) on the their validation set. 

{\noindent\bf{Experimental details.}}
We capture multiple video sequences at 1920×1080 resolution using a DJI Matrice 4T camera~\cite{dji_matrice4t} across various AI surveillance scenarios, including elevator, street, room, and shopping mall settings. 
For privacy protection, all captured facial data are anonymized. In the digital experiments, the adversarial patch is applied to all person targets in each image. The patch is placed at the top-center of the bounding box and scaled to cover no more than 30\% of the person's body area. To ensure fair comparison, our polygonal patch maintains the same area as prior square or rectangular patches. Patch training data include the INRIA~\cite{41-dalal2005histograms} and LLVIP~\cite{39-jia2021llvip} datasets, along with additional visible- and infrared-spectrum images collected from real-world scenes. In the physical experiments, we selected the top-10 adversarial patches with the highest attack success rates in the digital domain and fabricated them into XL-sized T-shirts (length 75\,cm). Each patch covers an area no larger than $50\times50$\,cm and is positioned at the center of the clothing. The effectiveness of the adversarial patches is evaluated using the Attack Success Rate (ASR), where the detection confidence threshold is set to 0.5. ASR is defined as:$\text{ASR} = \frac{N_{\text{clean}} - N_{\text{patch}}}{N_{\text{clean}}}$, where $N_{\text{clean}}$ and $N_{\text{patch}}$ denote the number of detected objects on clean and adversarial images, respectively.

 \begin{figure*}[t]
\centering
\includegraphics[width=1\linewidth]{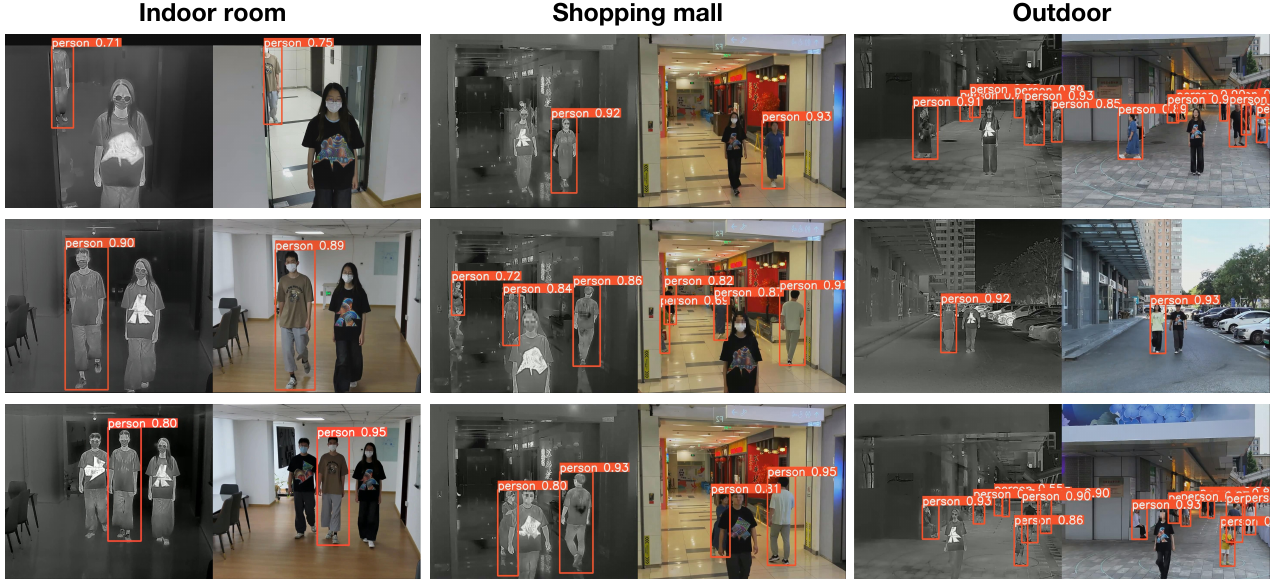}
\caption{\textbf{Real-world testing in diverse AI surveillance scenarios, including indoor room, shopping mall, and outdoor street.} This demonstrates that our adversarial method achieves effective dual-modal evasion on both visible and infrared detectors. Refer to Supplementary Materials for video demos.}
\vspace{-0.1in}
 \label{fig:multi-scenes}
\end{figure*}


 \begin{figure}[t]
\centering
\includegraphics[width=1.0\linewidth]{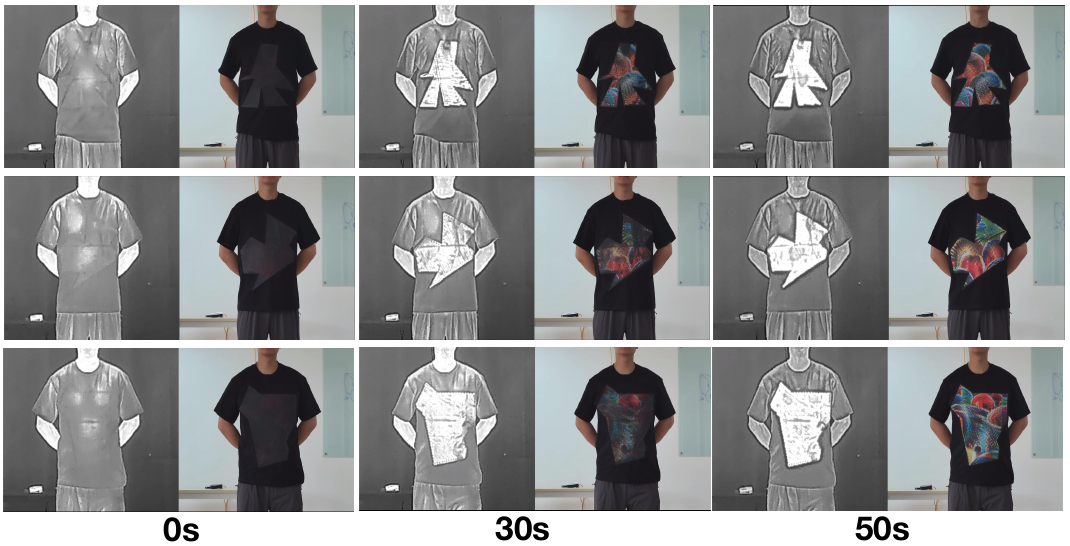}
\caption{\textbf{Fast activation of thermally activated adversarial clothing.} The hidden patches become revealed within 50~s.}
\vspace{-0.07in}
 \label{fig:patch-activation-time}
\end{figure}

 \begin{figure}[t]
\centering
\includegraphics[width=1\linewidth]{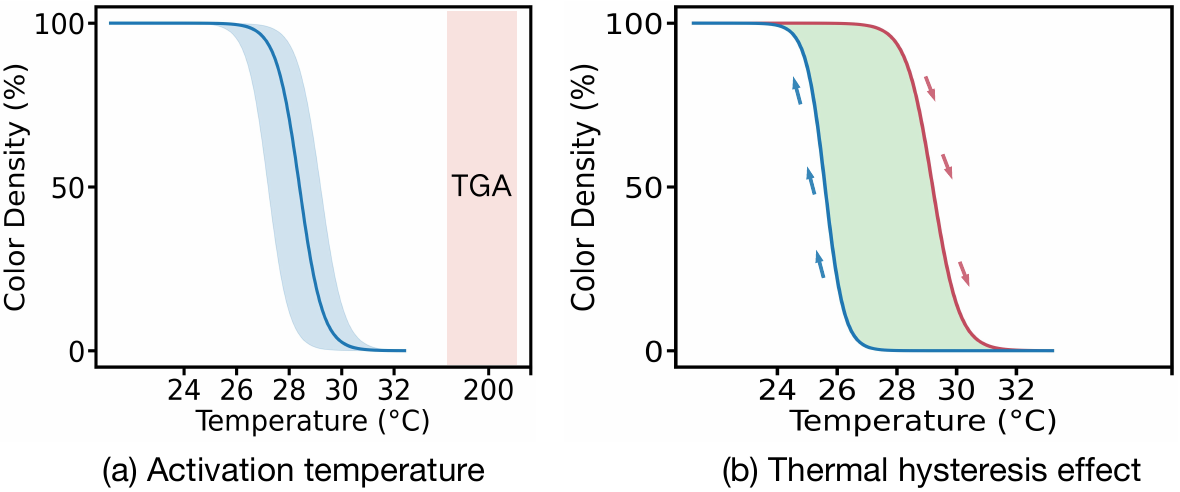}
\caption{\textbf{The activation temperature of the adversarial clothing.} (a) Dye color density vs. temperature: The dye undergoes a sharp transition from colored to transparent between 28-30 ºC. TGA indicates that the microcapsules lose protective capability above ~200 ºC. (b) Heating-cooling curves: The thermochromic material exhibits a thermal hysteresis of ~3 ºC.}
 \label{fig:Activation}
\end{figure}

\subsection{Main Results}
{\noindent\bf{Quantitative analysis.}} Figure~\ref{fig:data} compares the performance of various adversarial patch methods on the INRIA~\cite{41-dalal2005histograms}, PennyFudan~\cite{42-wang2007object}, FILR~\cite{40-flir_dataset} and LLVIP~\cite{39-jia2021llvip} Datasets. As can be seen, our dual-modal patch consistently achieves the highest attack success rate (ASR) across four mainstream detectors (i.e., Yolov3, Yolov5, FastRCNN, DETR), outperforming both visible patch methods (i.e., T-SEA~\cite{36-huang2023t}, AdvYOLO~\cite{24-thys2019fooling}, CAP~\cite{25-wei2024revisiting}, FDA~\cite{37-cheng2024full}, AdvTexture~\cite{30-hu2022adversarial}, AdvCloak~\cite{38-wu2020making}) and infrared patch methods (AdvIC31, AdvIB~\cite{32-hu2024adversarial}, AIP~\cite{33-wei2023physically}, HIC-IR~\cite{34-zhu2023hiding}, Bulb~\cite{35-zhu2024hiding}). For example, on the INRIA dataset under visible-spectrum detection, our patch attains ASRs of 54.4\% (YOLOv3), 49.9\% (YOLOv5), 46.8\% (Faster R-CNN), and 38.4\% (DETR), surpassing AdvYOLO (34.6\%, 31.4\%, 29.7\%, and 25.2\%, respectively). On the LLVIP dataset under infrared detection, our patch achieves 85.4\%, 88.6\%, 83.3\%, and 80.2\%, compared to AdvIB’s 72.7\%, 75.1\%, 70.3\%, and 65.8\%. Unlike prior square-shaped patches (e.g., CAP~\cite{25-wei2024revisiting}, AdvYOLO~\cite{24-thys2019fooling}), our design employs an algorithmically optimized polygonal geometry. This irregular shape enhances patch effectiveness against both visible- and infrared-spectrum detectors, allowing superior performance across diverse datasets and complex environments.

{\noindent\bf{Qualitative analysis.}} Figure~\ref{fig:multi-scenes} presents qualitative evaluations of our adversarial clothing under multiple real-world surveillance scenarios, including indoor rooms, shopping malls, and outdoor streets. Across all scenarios, our adversarial clothing reduces detection confidence and achieves dual-modal evasion across both visible and infrared detectors, demonstrating its effectiveness against complex AI surveillance. Figure~\ref{fig:multi-scenes}(a) illustrates an indoor surveillance scenario, including both room and elevator monitoring. As shown, individuals wearing either our adversarial clothing or ordinary clothing walk through the area. In both infrared and visible modes, those wearing the adversarial clothing are not detected by the detectors, whereas surrounding individuals are successfully detected with high confidence. This demonstrates the effectiveness of our adversarial clothing in indoor surveillance scenarios. 
Figure~\ref{fig:multi-scenes}(b) shows a shopping surveillance scenario with dense pedestrian traffic. Despite the crowded environment, the adversarial clothing remains effective: the wearer is not detected by the dual-modal detector, while other people in the scene are correctly identified. This demonstrates the robustness of our adversarial clothing in high-density pedestrian settings. 
Figure~\ref{fig:multi-scenes}(c) illustrate an outdoor street surveillance scenario. In this setting, the adversarial clothing continues to degrade detection across both infrared and visible modes: the wearer is missed by the dual-modal detector, while surrounding pedestrians are accurately recognized. This further confirms the robustness of our method in unconstrained outdoor environments.

\begin{figure*}[t]
\centering
\includegraphics[width=0.82\linewidth]{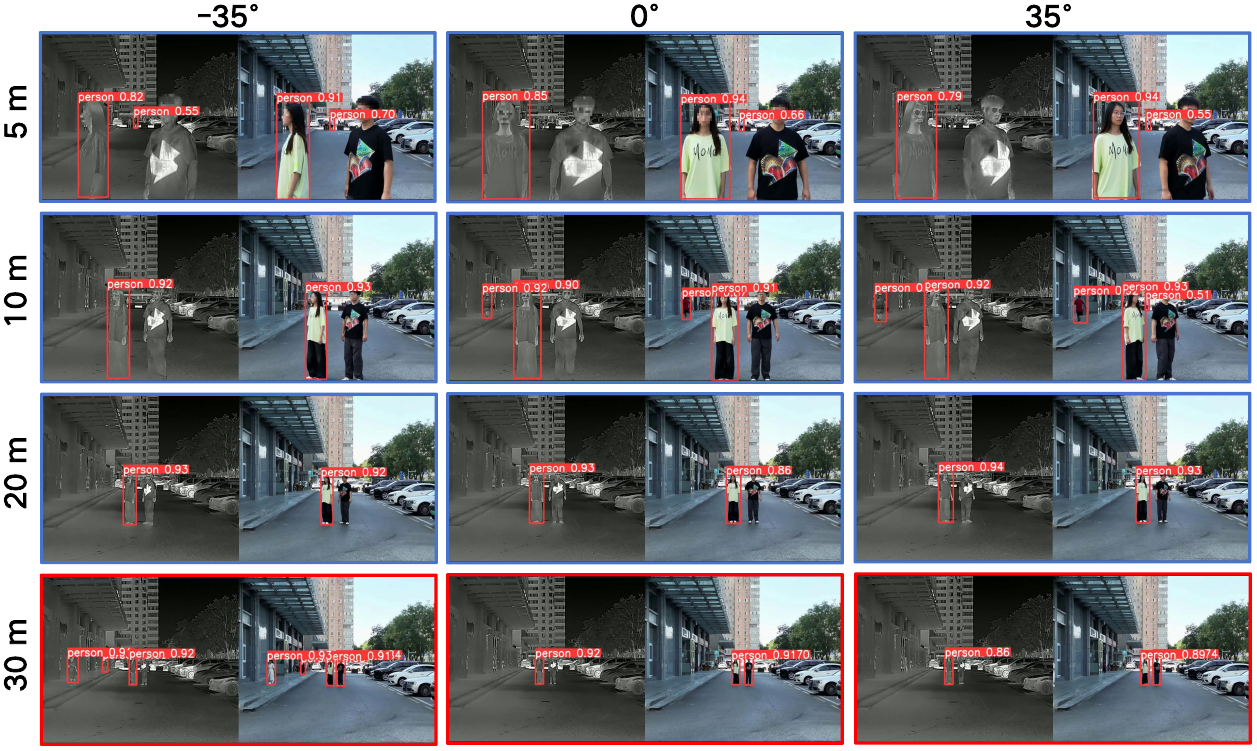}
\vspace{-0.1in}
\caption{\textbf{Real-world testing of patch effectiveness under diverse angles and distances.} It demonstrates that our clothing maintains attack effectiveness within varying physical conditions.}
 \label{fig:multi-angle}
\end{figure*}

\section{Ablation Study}

{\noindent\bf{Activation time.}}
As shown in Figure~\ref{fig:patch-activation-time}, we evaluated the response speed of the thermally activated adversarial clothing under controlled heating. Initially, the adversarial patterns remained invisible in both spectra; after 30 s of heating, the adversarial patterns started to emerge, and full activation was achieved within 50 s. This rapid response ensures reliable RGB texture activation for visible-spectrum deception while simultaneously altering thermal textures for infrared-spectrum deception, thereby achieving privacy protection within a short time window.
 
{\noindent\bf{Activation temperature.}}
Figure~\ref{fig:Activation}(a) exhibits the variation in dye color density as the temperature increases. The color density drops rapidly from 100\% (fully visible) to 0\% (nearly transparent) between 28-32 ºC, with complete transparency at ~32 ºC, at which point the adversarial patch becomes visible on the clothing. The pink region represents the Thermogravimetric Analysis (TGA) data, showing the microcapsules lose stability above ~200 ºC, beyond which the adversarial functionality can no longer be maintained. Figure~\ref{fig:Activation}(b) illustrates the thermal hysteresis effect of the thermochromic layer, where the transition temperatures during heating and cooling differ, forming a hysteresis loop (green shaded area). This behavior arises from the metastable phase state of the thermochromic dye, which is sensitive to thermal history.

{\noindent\bf{Effectiveness Boundary.}} We examine two key physical factors that affect the effectiveness of patch attacks: viewing angle and distance. 
Figure~\ref{fig:multi-angle} qualitatively analyzes the performance of our adversarial clothing under diverse angles and distances in real-world scenarios. The results demonstrate that our clothing maintains attack effectiveness within certain ranges. Specifically, at distances up to 30 m, the person wearing adversarial clothing is not detected by the visible-infrared dual-modal detectors, while surrounding individuals are detected. Beyond 30 m, the adversarial clothing becomes less effective against visible-spectrum detectors, but remains effective against infrared detectors. The degradation of the RGB patch arises is due to the patch becoming too small and blurred in the long distance, failing to perturb visual features of the person. Additionally, our adversarial clothing remains effective across a range of viewing angles within 35$^{\circ}$, indicating robustness against rotations.

\section{Conclusion}
In this paper, we have developed thermally activated adversarial clothing designed to counter AI-driven surveillance systems. The design integrates thermochromic dyes, flexible heating units, and adversarial patches to induce dynamic adversarial patterns on the clothing, enabling users to evade both visible and infrared AI detection. Unlike previous patch methods, our method conceals the adversarial patch within the clothing and allows users to actively trigger it in specific scenarios. By leveraging temperature as a control signal, the system simultaneously activates both RGB and infrared patches, thereby achieving dual-modal evasion in a controllable manner. Extensive physical experiments demonstrate that our design provide a practical pathway for enhancing personal privacy against AI-driven monitoring. 
{\noindent\bf{Acknowledges.}}
This work was supported in part by NSFC (62322113, 62376156), Shanghai Municipal Science and Technology Major Project (2025SHZDZX025G15), and the Foundation of the Chinese Academy of Military Science.

{
    \small
    \bibliographystyle{ieeenat_fullname}
    \bibliography{main}

@String(CVPR= {IEEE Conf. Comput. Vis. Pattern Recog.})

@String(ICCV= {Int. Conf. Comput. Vis.})

@String(ECCV= {Eur. Conf. Comput. Vis.})

@String(ACCV  = {ACCV})

@String(AAAI = {AAAI})

@String(CVPRW= {IEEE Conf. Comput. Vis. Pattern Recog. Worksh.})

@String(CVPR  = {CVPR})

@String(ICCV  = {ICCV})

@String(ECCV  = {ECCV})

@String(CVPRW= {CVPRW})

@article{1-hussain2024ai,
  title={AI-driven behavior biometrics framework for robust human activity recognition in surveillance systems},
  author={Hussain, Altaf and Khan, Samee Ullah and Khan, Noman and Shabaz, Mohammad and Baik, Sung Wook},
  journal={Engineering Applications of Artificial Intelligence},
  volume={127},
  pages={107218},
  year={2024}
}

@article{2-kalluri2025computer,
  title={Computer-vision research powers surveillance technology},
  author={Kalluri, Pratyusha Ria and Agnew, William and Cheng, Myra and Owens, Kentrell and Soldaini, Luca and Birhane, Abeba},
  journal={Nature},
  volume={643},
  number={8070},
  pages={73--79},
  year={2025}}

@incollection{3-zuboff2023surveillance,
  author       = {Zuboff, Shoshana},
  title        = {The Age of Surveillance Capitalism},
  booktitle    = {Social Theory Re-Wired},
  pages        = {203-213},
  publisher    = {Routledge},
  year         = {2023}
}

@article{4-maslej2025artificial,
  title={Artificial intelligence index report 2025},
  author={Maslej, Nestor and Fattorini, Loredana and Perrault, Raymond and Gil, Yolanda and Parli, Vanessa and Kariuki, Njenga and Capstick, Emily and Reuel, Anka and Brynjolfsson, Erik and Etchemendy, John and others},
  journal={arXiv:2504.07139},
  year={2025}
}

@book{5-feldstein2019global,
  title={The global expansion of AI surveillance},
  author={Feldstein, Steven},
  volume={17},
  number={9},
  year={2019},
  publisher={Carnegie Endowment for International Peace Washington, DC}
}

@article{6-ardabili2023understanding,
  title={Understanding policy and technical aspects of ai-enabled smart video surveillance to address public safety},
  author={Ardabili, Babak Rahimi and Pazho, Armin Danesh and Noghre, Ghazal Alinezhad and Neff, Christopher and Bhaskararayuni, Sai Datta and Ravindran, Arun and Reid, Shannon and Tabkhi, Hamed},
  journal={Computational Urban Science},
  volume={3},
  number={1},
  pages={21},
  year={2023}}

@book{8-browne2015dark,
  title={Dark matters: On the surveillance of blackness},
  author={Browne, Simone},
  year={2015},
  publisher={Duke University Press}
}

@online{9-chang2022countermeasures,
  author       = {Chang, M.},
  title        = {Countermeasures: The Need for New Legislation to Govern Biometric Technologies in the UK},
  year         = {2022},
  url          = {https://www.adalovelaceinstitute.org/report/countermeasures/},
  note         = {Ada Lovelace Institute Report}
}

@article{11-waelen2024ethics,
  title={The ethics of computer vision: an overview in terms of power},
  author={Waelen, Rosalie A},
  journal={AI and Ethics},
  volume={4},
  number={2},
  pages={353-362},
  year={2024}}

@article{12-van2020ethical,
  title={The ethical questions that haunt facial-recognition research},
  author={Van Noorden, Richard},
  journal={Nature},
  volume={587},
  number={7834},
  pages={354--359},
  year={2020}
}

@online{13-perrigo2022clearview,
  author       = {Perrigo, Billy},
  title        = {An AI Company Scraped Billions of Photos for Facial Recognition},
  year         = {2022},
  journal      = {TIME},
  url          = {https://time.com/6182177/clearview-ai-regulators-uk/}}

@article{14-barkane2022questioning,
  title={Questioning the EU proposal for an Artificial Intelligence Act: The need for prohibitions and a stricter approach to biometric surveillance},
  author={Barkane, Irena},
  journal={Information Polity},
  volume={27},
  number={2},
  pages={147-162},
  year={2022}}

@online{15-eu2024aiact,
  author       = {{European Parliament and the Council}},
  title        = {Regulation (EU) 2024/1689},
  year         = {2024},
  url          = {https://artificialintelligenceact.eu/the-act/},
  note         = {Official EU Legislation Portal}
}

@online{16-bird2025aiguide,
  author       = {{Bird \& Bird LLP}},
  title        = {European Union Artificial Intelligence Act: A Guide},
  year         = {2024},
  url          = {https://www.twobirds.com/-/media/new-website-content/pdfs/capabilities/artificial-intelligence/european-union-artificial-intelligence-act-guide.pdf},
  note         = {https://www.twobirds.com/-/media/new-website-content/pdfs/capabilities/artificial-intelligence/european-union-artificial-intelligence-act-guide.pdf}
}

@online{17-eprs2024legislation,
  author       = {{European Parliamentary Research Service}},
  title        = {EU Legislation in Progress},
  year         = {2024},
  url          = {https://epthinktank.eu/eu-legislation-in-progress/},
  note         = {https://epthinktank.eu/eu-legislation-in-progress/}
}

@online{18-davis2024stateai,
  author       = {{Davis+Gilbert LLP}},
  title        = {Utah, Colorado and Other States Lead Groundbreaking AI Legislation in the U.S.},
  year         = {2024}}

@inproceedings{19-hingun2023reap,
  title={REAP: A large-scale realistic adversarial patch benchmark},
  author={Hingun, Nabeel and Sitawarin, Chawin and Li, Jerry and Wagner, David},
  booktitle={Proceedings of the IEEE/CVF International Conference on Computer Vision (ICCV)},
  year={2023}
}

@article{20-hu2025two,
  title={Two-stage optimized unified adversarial patch for attacking visible-infrared cross-modal detectors in the physical world},
  author={Hu, Chengyin and Shi, Weiwen and Yao, Wen and Jiang, Tingsong and Tian, Ling and Li, Wen},
  journal={Applied Soft Computing},
  volume={171},
  pages={112818},
  year={2025}
}

@inproceedings{21-wei2023unified,
  title={Unified adversarial patch for cross-modal attacks in the physical world},
  author={Wei, Xingxing and Huang, Yao and Sun, Yitong and Yu, Jie},
  booktitle={Proceedings of the IEEE/CVF International Conference on Computer Vision (ICCV)},
  pages={4445--4454},
  year={2023}
}

@inproceedings{22-long2025cdupatch,
  title={Cdupatch: Color-driven universal adversarial patch attack for dual-modal visible-infrared detectors},
  author={Long, Jiahuan and Yao, Wen and Jiang, Tingsong and Hou, Jiacheng and Jia, Shuai and Wu, Junqi and Zhang, Xiaoya and Zheng, Xiaohu and Ma, Chao},
  booktitle={Proceedings of the 33rd ACM International Conference on Multimedia (ACM MM)},
  year={2025}
}

@article{23-brown2017adversarial,
  title={Adversarial patch},
  author={Brown, Tom B and Man{\'e}, Dandelion and Roy, Aurko and Abadi, Mart{\'\i}n and Gilmer, Justin},
  journal={arXiv:1712.09665},
  year={2017}
}

@inproceedings{24-thys2019fooling,
  title={Fooling automated surveillance cameras: adversarial patches to attack person detection},
  author={Thys, Simen and Van Ranst, Wiebe and Goedem{\'e}, Toon},
  booktitle={Proceedings of the IEEE/CVF conference on computer vision and pattern recognition workshops (CVPRW)},
  year={2019}
}

@article{25-wei2024revisiting,
  title={Revisiting adversarial patches for designing camera-agnostic attacks against person detection},
  author={Wei, Hui and Wang, Zhixiang and Zhang, Kewei and Hou, Jiaqi and Liu, Yuanwei and Tang, Hao and Wang, Zheng},
  journal={Advances in Neural Information Processing Systems (NeurIPS)},
  year={2024}
}

@article{26-liu2018dpatch,
  title={Dpatch: An adversarial patch attack on object detectors},
  author={Liu, Xin and Yang, Huanrui and Liu, Ziwei and Song, Linghao and Li, Hai and Chen, Yiran},
  journal={arXiv:1806.02299},
  year={2018}
}

@inproceedings{27-long2024papmot,
  title={PapMOT: Exploring Adversarial Patch Attack Against Multiple Object Tracking},
  author={Long, Jiahuan and Jiang, Tingsong and Yao, Wen and Jia, Shuai and Zhang, Weijia and Zhou, Weien and Ma, Chao and Chen, Xiaoqian},
  booktitle={European Conference on Computer Vision (ECCV)},
  year={2024}}

@inproceedings{28-xu2020adversarial,
  title={Adversarial t-shirt! evading person detectors in a physical world},
  author={Xu, Kaidi and Zhang, Gaoyuan and Liu, Sijia and Fan, Quanfu and Sun, Mengshu and Chen, Hongge and Chen, Pin-Yu and Wang, Yanzhi and Lin, Xue},
  booktitle={European conference on computer vision (ECCV)},
  year={2020}}

@inproceedings{30-hu2022adversarial,
  title={Adversarial texture for fooling person detectors in the physical world},
  author={Hu, Zhanhao and Huang, Siyuan and Zhu, Xiaopei and Sun, Fuchun and Zhang, Bo and Hu, Xiaolin},
  booktitle={Proceedings of the IEEE/CVF conference on computer vision and pattern recognition (CVPR)},
  year={2022}
}

@article{31-hu2024adversarial,
  title={Adversarial infrared curves: An attack on infrared pedestrian detectors in the physical world},
  author={Hu, Chengyin and Shi, Weiwen and Yao, Wen and Jiang, Tingsong and Tian, Ling and Chen, Xiaoqian and Li, Wen},
  journal={Neural networks},
  volume={178},
  pages={106459},
  year={2024}
}

@article{32-hu2024adversarial,
  title={Adversarial infrared blocks: A multi-view black-box attack to thermal infrared detectors in physical world},
  author={Hu, Chengyin and Shi, Weiwen and Jiang, Tingsong and Yao, Wen and Tian, Ling and Chen, Xiaoqian and Zhou, Jingzhi and Li, Wen},
  journal={Neural Networks},
  volume={175},
  pages={106310},
  year={2024}
}

@inproceedings{33-wei2023physically,
  title={Physically adversarial infrared patches with learnable shapes and locations},
  author={Wei, Xingxing and Yu, Jie and Huang, Yao},
  booktitle={Proceedings of the IEEE/CVF conference on computer vision and pattern recognition (CVPR)},
  year={2023}
}

@article{34-zhu2023hiding,
  title={Hiding from infrared detectors in real world with adversarial clothes},
  author={Zhu, Xiaopei and Hu, Zhanhao and Huang, Siyuan and Li, Jianmin and Hu, Xiaolin and Wang, Zheyao},
  journal={Applied Intelligence},
  volume={53},
  number={23},
  pages={29537-29555},
  year={2023}
}

@article{35-zhu2024hiding,
  title={Hiding from thermal imaging pedestrian detectors in the physical world},
  author={Zhu, Xiaopei and Li, Xiao and Li, Jianmin and Wang, Zheyao and Hu, Xiaolin},
  journal={Neurocomputing},
  volume={564},
  pages={126923},
  year={2024}
}

@inproceedings{36-huang2023t,
  title={T-sea: Transfer-based self-ensemble attack on object detection},
  author={Huang, Hao and Chen, Ziyan and Chen, Huanran and Wang, Yongtao and Zhang, Kevin},
  booktitle={Proceedings of the IEEE/CVF conference on computer vision and pattern recognition (CVPR)},
  pages={20514-20523},
  year={2023}
}

@article{37-cheng2024full,
  title={Full-distance evasion of pedestrian detectors in the physical world},
  author={Cheng, Zhi and Hu, Zhanhao and Liu, Yuqiu and Li, Jianmin and Su, Hang and Hu, Xiaolin},
  journal={Advances in Neural Information Processing Systems (NeurIPS)},
  year={2024}
}

@inproceedings{38-wu2020making,
  title={Making an invisibility cloak: Real world adversarial attacks on object detectors},
  author={Wu, Zuxuan and Lim, Ser-Nam and Davis, Larry S and Goldstein, Tom},
  booktitle={European Conference on Computer Vision (ECCV)},
  year={2020}
}

@inproceedings{39-jia2021llvip,
  title={LLVIP: A visible-infrared paired dataset for low-light vision},
  author={Jia, Xinyu and Zhu, Chuang and Li, Minzhen and Tang, Wenqi and Zhou, Wenli},
  booktitle={Proceedings of the IEEE/CVF international conference on computer vision (ICCV)},
  year={2021}
}

@misc{40-flir_dataset,
  title        = {https://oem.flir.com/},
  author       = {{FLIR Thermal Datasets}}

}

@inproceedings{41-dalal2005histograms,
  title={Histograms of oriented gradients for human detection},
  author={Dalal, Navneet and Triggs, Bill},
  booktitle={Proceedings of the IEEE/CVF conference on computer vision and pattern recognition (CVPR)},
  volume={1},
  pages={886--893},
  year={2005}
}

@inproceedings{42-wang2007object,
  title={Object detection combining recognition and segmentation},
  author={Wang, Liming and Shi, Jianbo and Song, Gang and Shen, I-fan},
  booktitle={Asian conference on computer vision (ACCV)},
  year={2007}
}

@article{43-redmon2018yolov3,
  title={Yolov3: An incremental improvement},
  author={Redmon, Joseph and Farhadi, Ali},
  journal={arXiv:1804.02767},
  year={2018}
}

@misc{44-jocher2020yolov5,
  author       = {Jocher, Glenn and others},
  title        = {YOLOv5 by Ultralytics, {https://doi.org/10.5281/zenodo.4154370}},
  year         = {2020}}

@article{45-ren2015faster,
  title={Faster r-cnn: Towards real-time object detection with region proposal networks},
  author={Ren, Shaoqing and He, Kaiming and Girshick, Ross and Sun, Jian},
  journal={Advances in neural information processing systems (NeurIPS)},
  year={2015}
}

@inproceedings{46-carion2020end,
  title={End-to-end object detection with transformers},
  author={Carion, Nicolas and Massa, Francisco and Synnaeve, Gabriel and Usunier, Nicolas and Kirillov, Alexander and Zagoruyko, Sergey},
  booktitle={European conference on computer vision (ECCV)},
  year={2020}
}

@inproceedings{47-lin2014microsoft,
  title={Microsoft coco: Common objects in context},
  author={Lin, Tsung-Yi and Maire, Michael and Belongie, Serge and Hays, James and Perona, Pietro and Ramanan, Deva and Doll{\'a}r, Piotr and Zitnick, C Lawrence},
  booktitle={European conference on computer vision (ECCV)},
  year={2014}
}

@inproceedings{48-athalye2018synthesizing,
  title={Synthesizing robust adversarial examples},
  author={Athalye, Anish and Engstrom, Logan and Ilyas, Andrew and Kwok, Kevin},
  booktitle={International conference on machine learning (ICML)},
  year={2018}
}

@misc{49-adversarialfashion,
  title          = {https://adversarialfashion.com},
  author        = {Adversarial Fashion}
}

@misc{50-capable2025clothing,
  title          = {https://www.capable.design},
  author        = {A Clothing for Data Privacy}
}

@article{51-liu2020reversible,
  title={Reversible nontoxic thermochromic microcapsules},
  author={Liu, Bingxin and Rasines Mazo, Alicia and Gurr, Paul A and Qiao, Greg G},
  journal={ACS applied materials \& interfaces},
  volume={12},
  number={8},
  pages={9782--9789},
  year={2020}
}

@article{52-liu2023high,
  title={High-Temperature, Reversible, and Robust Thermochromic Fluorescence Based on Rb2MnBr4 (H2O) 2 for Anti-Counterfeiting},
  author={Liu, Yang and Liu, Gaoyu and Wu, Ye and Cai, Wenbing and Wang, Yue and Zhang, Shengli and Zeng, Haibo and Li, Xiaoming},
  journal={Advanced Materials},
  volume={35},
  number={35},
  pages={2301914},
  year={2023}
}

@article{54-tang2024electrothermochromic,
  title={Electrothermochromic Fabrics with a Single-Layer Functional Coating Based on Silver Nanowires/Thermochromic Microcapsules/Waterborne Polyurethane Paints},
  author={Tang, Jingli and Wang, Yichao and He, Mengjuan and Huang, Liqian and Wang, Xueli and Yu, Jianyong},
  journal={ACS Applied Materials \& Interfaces},
  volume={16},
  number={51},
  pages={70963--70972},
  year={2024}
}

@misc{dji_matrice4t,
  title        = {https://enterprise.dji.com/matrice-4-series},
  author          = {DJI Matrice 4T}
}

@inproceedings{long2025robust,
  title={Robust SAM: On the Adversarial Robustness of Vision Foundation Models},
  author={Long, Jiahuan and Xu, Zhengqin and Jiang, Tingsong and Yao, Wen and Jia, Shuai and Ma, Chao and Chen, Xiaoqian},
  booktitle={Proceedings of the AAAI Conference on Artificial Intelligence (AAAI)},
  year={2025}
}

@article{liu2025lighting,
  title={When Lighting Deceives: Exposing Vision-Language Models' Illumination Vulnerability Through Illumination Transformation Attack},
  author={Liu, Hanqing and Ruan, Shouwei and Huang, Yao and Zhao, Shiji and Wei, Xingxing},
  journal={arXiv:2503.06903},
  year={2025}
}

@inproceedings{jia2021iou,
  title={Iou attack: Towards temporally coherent black-box adversarial attack for visual object tracking},
  author={Jia, Shuai and Song, Yibing and Ma, Chao and Yang, Xiaokang},
  booktitle={Proceedings of the IEEE/CVF Conference on Computer Vision and Pattern Recognition},
  year={2021}
}

@inproceedings{jia2022exploring,
  title={Exploring frequency adversarial attacks for face forgery detection},
  author={Jia, Shuai and Ma, Chao and Yao, Taiping and Yin, Bangjie and Ding, Shouhong and Yang, Xiaokang},
  booktitle={Proceedings of the IEEE/CVF Conference on Computer Vision and Pattern Recognition},
  year={2022}
}

@inproceedings{zhu2024infrared,
  title={Infrared adversarial car stickers},
  author={Zhu, Xiaopei and Liu, Yuqiu and Hu, Zhanhao and Li, Jianmin and Hu, Xiaolin},
  booktitle={Proceedings of the IEEE/CVF Conference on Computer Vision and Pattern Recognition (CVPR)},
  year={2024}
}

@inproceedings{huang2024towards,
  title={Towards transferable targeted 3d adversarial attack in the physical world},
  author={Huang, Yao and Dong, Yinpeng and Ruan, Shouwei and Yang, Xiao and Su, Hang and Wei, Xingxing},
  booktitle={Proceedings of the IEEE/CVF Conference on Computer Vision and Pattern Recognition (CVPR)},
  year={2024}
}

@article{zhu2025physical,
  title={Physical Adversarial Examples for Person Detectors in Thermal Images Based on 3D Modeling},
  author={Zhu, Xiaopei and Huang, Siyuan and Hu, Zhanhao and Li, Jianmin and Zhu, Jun and Hu, Xiaolin},
  journal={IEEE Transactions on Pattern Analysis and Machine Intelligence},
  year={2025},
  publisher={IEEE}
}

@article{liu2025eva,
  title={Eva-VLA: Evaluating Vision-Language-Action Models' Robustness Under Real-World Physical Variations},
  author={Liu, Hanqing and Long, Jiahuan and Wu, Junqi and Hou, Jiacheng and Tang, Huili and Jiang, Tingsong and Zhou, Weien and Yao, Wen},
  journal={arXiv:2509.18953},
  year={2025}
}

@inproceedings{hu2023physically,
  title={Physically realizable natural-looking clothing textures evade person detectors via 3d modeling},
  author={Hu, Zhanhao and Chu, Wenda and Zhu, Xiaopei and Zhang, Hui and Zhang, Bo and Hu, Xiaolin},
  booktitle={Proceedings of the IEEE/CVF Conference on Computer Vision and Pattern Recognition},
  year={2023}
}

@article{guo2025physpatch,
  title={Physpatch: A physically realizable and transferable adversarial patch attack for multimodal large language models-based autonomous driving systems},
  author={Guo, Qi and Jia, Xiaojun and Pang, Shanmin and Qin, Simeng and Wang, Lin and Jia, Ju and Liu, Yang and Guo, Qing},
  journal={arXiv:2508.05167},
  year={2025}
}

@inproceedings{hingun2023reap,
  title={REAP: A large-scale realistic adversarial patch benchmark},
  author={Hingun, Nabeel and Sitawarin, Chawin and Li, Jerry and Wagner, David},
  booktitle={Proceedings of the IEEE/CVF International Conference on Computer Vision (ICCV)},
  year={2023}
}

@inproceedings{chen2022shape,
  title={Shape matters: deformable patch attack},
  author={Chen, Zhaoyu and Li, Bo and Wu, Shuang and Xu, Jianghe and Ding, Shouhong and Zhang, Wenqiang},
  booktitle={European conference on computer vision (ECCV)},
  year={2022}
}

@inproceedings{zheng2024physical,
  title={Physical 3D adversarial attacks against monocular depth estimation in autonomous driving},
  author={Zheng, Junhao and Lin, Chenhao and Sun, Jiahao and Zhao, Zhengyu and Li, Qian and Shen, Chao},
  booktitle={Proceedings of the IEEE/CVF Conference on Computer Vision and Pattern Recognition (CVPR)},
  year={2024}
}

@inproceedings{zheng2025revisiting,
  title={Revisiting Adversarial Patch Defenses on Object Detectors: Unified Evaluation, Large-Scale Dataset, and New Insights},
  author={Zheng, Junhao and Sun, Jiahao and Lin, Chenhao and Zhao, Zhengyu and Ma, Chen and Zhang, Chong and Wang, Cong and Wang, Qian and Shen, Chao},
  booktitle={Proceedings of the IEEE/CVF International Conference on Computer Vision (ICCV)},
  year={2025}
}

@article{wei2022adversarial,
  title={Adversarial sticker: A stealthy attack method in the physical world},
  author={Wei, Xingxing and Guo, Ying and Yu, Jie},
  journal={IEEE Transactions on Pattern Analysis and Machine Intelligence},
  volume={45},
  number={3},
  pages={2711--2725},
  year={2022}
}

@article{wei2024infrared,
  title={Infrared adversarial patches with learnable shapes and locations in the physical world},
  author={Wei, Xingxing and Yu, Jie and Huang, Yao},
  journal={International Journal of Computer Vision},
  volume={132},
  number={6},
  pages={1928--1944},
  year={2024}
}

@inproceedings{wang2021dual,
  title={Dual attention suppression attack: Generate adversarial camouflage in physical world},
  author={Wang, Jiakai and Liu, Aishan and Yin, Zixin and Liu, Shunchang and Tang, Shiyu and Liu, Xianglong},
  booktitle={Proceedings of the IEEE/CVF conference on computer vision and pattern recognition (CVPR)},
  year={2021}
}

@article{wang2021universal,
  title={Universal adversarial patch attack for automatic checkout using perceptual and attentional bias},
  author={Wang, Jiakai and Liu, Aishan and Bai, Xiao and Liu, Xianglong},
  journal={IEEE Transactions on Image Processing},
  volume={31},
  pages={598--611},
  year={2021}
}

@inproceedings{liu2019perceptual,
  title={Perceptual-sensitive gan for generating adversarial patches},
  author={Liu, Aishan and Liu, Xianglong and Fan, Jiaxin and Ma, Yuqing and Zhang, Anlan and Xie, Huiyuan and Tao, Dacheng},
  booktitle={Proceedings of the AAAI conference on artificial intelligence (AAAI)},
  year={2019}
}

@inproceedings{liu2020bias,
  title={Bias-based universal adversarial patch attack for automatic check-out},
  author={Liu, Aishan and Wang, Jiakai and Liu, Xianglong and Cao, Bowen and Zhang, Chongzhi and Yu, Hang},
  booktitle={European conference on computer vision (ECCV)},
  year={2020}
}
}


\end{document}